\pdfoutput=1

\documentclass[11pt]{article}

\usepackage[preprint]{acl}

\usepackage{times}
\usepackage{latexsym}

\usepackage[T1]{fontenc}

\usepackage[utf8]{inputenc}

\usepackage{microtype}

\usepackage{inconsolata}

\usepackage{graphicx}

\usepackage{subcaption}

\usepackage{booktabs}

\usepackage[many]{tcolorbox}

\title{Can You Trust LLM Judgments? Reliability of LLM-as-a-Judge}

\author{Kayla Schroeder \\
  Department of Statistics \\
  Northwestern University \\
  \texttt{kaylaschroeder2026@u.northwestern.edu} \\\And
  Zach Wood-Doughty \\
  Department of Computer Science \\
  Northwestern University \\
  \texttt{zach@northwestern.edu} \\}

\begin{document}
\maketitle
\begin{abstract}

Large Language Models (LLMs) have become increasingly powerful and ubiquitous, but their stochastic nature poses challenges to the reliability of their outputs. While deterministic settings can improve consistency, they do not guarantee reliability, as a single sample from the model's probability distribution can still be misleading. Building upon the concept of LLM-as-a-judge, we introduce a novel framework for rigorously evaluating the reliability of LLM judgments, leveraging McDonald's omega. We evaluate the reliability of LLMs when judging the outputs of other LLMs on standard single-turn and multi-turn benchmarks, simultaneously investigating the impact of temperature on reliability. By analyzing these results, we demonstrate the limitations of fixed randomness and the importance of considering multiple samples, which we show has significant implications for downstream applications. Our findings highlight the need for a nuanced understanding of LLM reliability and the potential risks associated with over-reliance on single-shot evaluations. This work provides a crucial step towards building more trustworthy and reliable LLM-based systems and applications.

\end{abstract}

\section{Introduction}
\label{sec:intro}

In recent years, Large Language Models (LLMs) have experienced rapid advancements and widespread adoption, with applications ranging from marketing to biotechnology to music \citep{gozalo2023survey}. As reliance upon LLMs grows, ensuring the quality and trustworthiness of their outputs becomes crucial. We often think about LLM quality in terms of accuracy (how often the LLM is correct). However, equally important are reliability, confidence, and calibration; these three distinct concepts are intertwined and essential for building genuine trust in LLM systems.

LLMs can express how certain, or 'confident', they are about their responses. Calibration measures how well this confidence aligns with the LLM's actual accuracy \citep{kotelanski2023methods, spiess2024calibration}. A perfectly calibrated LLM that says it's 90\% confident should be correct about 9 out of 10 times. However, even a perfectly calibrated LLM can be unreliable.

Reliability refers to the consistency of an LLM's outputs. A reliable weather forecasting LLM, for example, should give similar predictions for the same day, even if those predictions aren't always accurate. An LLM that sometimes predicts sunshine and sometimes a downpour for the same day is unreliable, even if it's well-calibrated and highly confident. This kind of consistency is what we aim to capture with internal consistency reliability.  A reliable LLM can be consistently wrong, but it's still reliable because its judgments are stable.

Why is reliability so important? Consider a scenario where an LLM is tasked with evaluating the quality of a medical imaging LLM used to diagnose tumors. Even if the LLM is perfectly calibrated (its confidence accurately reflects its diagnostic ability), if it gives different evaluations for the same AI on different runs, even with different, but accurate, confidence scores, doctors won't know which evaluation to trust. The inconsistency makes the LLM unreliable and ultimately untrustworthy, regardless of its calibration and confidence.

A singular output from an LLM is the result of a draw from a probabilistic distribution \citep{vaswani2017attention}. Parameters that vary the resulting output, namely temperature and top-k, are increased to obtain more novel or "creative" results - a desirable quality of LLMs \citep{peeperkorn2024temperature}. However, this inherent stochastic nature raises concerns about individual outputs. Given that a single output from an LLM is a representation of only a single draw from the model's distribution, trustworthiness of an LLM output is of utmost importance \citep{mittelstadt2023protect}. Confidence and calibration are commonly used within NLP to understand this variability \citep{lin2023generating, tian2023just}, but they don't address reliability directly.

Many research works aim to circumvent this issue by setting a fixed seed and using deterministic settings for the temperature and top-k parameters \citep{ouyang2023llm, wei2024systematic, atil2024llm}. These studies argue that if an LLM consistently produces the same output under these conditions, it can be considered reliable. However, consistent replication does not guarantee the reliability of the generated text. 

Even with deterministic settings, a single LLM output remains a sample from the model's probability distribution, subject to inherent randomness. This results in "fixed randomness," which can lead to significant limitations \citep{hellrich2016bad}. Consider our prior example of an LLM evaluator of a medical imaging LLM for tumor diagnosis. If the evaluating LLM, due to its own inherent randomness, consistently fails to identify a critical weakness in the target LLM in a single evaluation run, grave repercussions can arise. Eradicating fixed randomness is crucial to ensure model consistency and reduce the risk of these systematic errors. Internal consistency reliability addresses this problem by assessing how consistently the LLM applies its evaluation criteria across multiple evaluations, rather than relying on a single, potentially flawed output.

The ability to trust LLM outputs is of further importance as LLMs are increasingly used as evaluators \citep{desmond2024evalullm, chan2023chateval}. The LLM-as-a-judge paradigm, coined by \citet{zheng2023judging}, relies upon the internal consistency reliability of LLM judges. Widespread utilization of LLMs as judges has garnered concern. The ACL Rolling Review Reviewer Guidelines highlight this issue, voicing explicit concern about LLM-only evaluation without validation by listing the methodological issue "If LLMs are used as automated evaluators, is their reliability in this context sufficiently validated?"\footnote{As of December 2024 at \url{https://aclrollingreview.org/reviewerguidelines}}. This underscores the critical need for trust in LLM outputs and their reliability. Reliability here, however, is not explicitly defined, and inconsistencies in how reliability of LLMs should be quantified are present throughout the literature.

While some research has focused on determining LLM reliability by comparing outputs to ground truth labels \citep{zhou2024llm, fu2023large}, this approach is often limited by the availability of accurate and comprehensive ground truth data. As LLMs are increasingly employed in complex and subjective domains, such as evaluating a medical imaging LLM for tumor diagnosis where even expert human diagnoses can vary, establishing a reliable ground truth becomes increasingly challenging. In current practice, LLM judgment models are often treated as raters, and inter-rater reliability is calculated on their outputs to assess reliability \citep{kollitsch2024does, wang2024prompt}. This metric, as we will show, is an insufficient quantification of reliability as the metric does not take into account the additional complexity of the LLM judge being a model itself and containing randomness.

The unique characteristics of LLMs as evaluators necessitate a more nuanced approach to assessing their reliability. We define reliability in this context as internal consistency reliability, a well-established statistical concept and the most common type of reliability employed across domains \citet{henson2001understanding}. While internal consistency is a well-established concept, its application to the LLM-as-a-judge paradigm, particularly with a focus on seed variation, has been largely overlooked. Internal consistency reliability measures how well different parts of an evaluation (individual LLM judgments) measure the same underlying construct (the LLM's "true" judgment). Internal consistency reliability ensures that the LLM applies its evaluation criteria consistently across various inputs and conditions. Metrics like Cronbach's alpha and McDonald's omega  provide quantifiable measures of this internal consistency, allowing us to assess the reliability of an LLM's evaluations by examining the consistency of its responses across a range of prompts or tasks \citep{cronbach1951coefficient, mcdonald2013test, malkewitz2023estimating}. 

We introduce a novel framework that leverages McDonald's omega to address a critical gap in current LLM evaluation methodologies by rigorously quantifying the stability and consistency of LLM judgments across multiple replicated evaluations of the same responses. Critically, this work pioneers the systematic investigation of the variability of an LLM evaluator's judgments solely by varying the random seed, holding all other parameters constant. We assess how consistently LLMs apply their evaluation criteria across replications (i.e., multiple independent evaluations of the same responses with fixed parameters), investigating the impact of diverse question formats and varying difficulty levels using established benchmarks (BBH, SQuAD, MT-Bench). Our findings provide novel insights into the sensitivity of LLM judgments, revealing potential limitations in their reliability and highlighting the importance of considering this variability in high-stakes applications. \footnote{The code for the framework and analyses can be accessed at \href{https://github.com/kaylaschroeder/llm_reliability}{https://github.com/kaylaschroeder/llm\_reliability}.}

\section{Related Work}

Reliability is a cornerstone of NLP research, particularly when human judgment is involved. Studies utilizing human labelers have long emphasized inter-annotator agreement \citep{bhowmick2008agreement,nowak2010reliable,amidei2019agreement}. Similarly, topic modeling has extensively explored reliability through various methods, including similarity measures, statistical techniques, and domain expertise \citep{rieger2024ldaprototype, schroeder2024reliability, chuang2015topiccheck}.

Internal consistency reliability itself, especially Cronbach's alpha, has been heavily utilized as an assessor of question quality in LLM contexts \citep{shang2024trusting, chan2023assessing, biri2023assessing}. \citet{bhandari2024evaluating} uses Cronbach's alpha to compare LLM-generated questions with human-authored ones; \citet{shi2023llm} uses Cronbach's alpha to assess the reliability of questions designed to evaluate LLMs. However, these studies focus on the reliability of questions, not the reliability of LLM judgments within the LLM-as-a-judge paradigm, a key distinction in our work.

The concept of reliability appears widely in the LLM literature, but is not consistently defined. Some studies equate reliability with human-model agreement, requiring costly human evaluations \citep{sun2024parrot, zhang2024can}. Others rely on accuracy metrics, which necessitate ground truth labels \citep{zheng2024response, macherla2023mddial, wei2024systematic}. As the complexity and subjectivity of judgment tasks increase, obtaining reliable ground truth becomes increasingly challenging. Semantic consistency has also been proposed as a reliability measure \citep{govil2024cobias, raj2023semantic}, but this approach lacks a strong theoretical foundation and has potential pitfalls, as discussed in \citet{schroeder2024reliability}. Many studies also use the term "reliability" without any rigorous definition \citep{tan2024peer, liao2023automatic, gupta2024llm}. This lack of a clear and consistent definition highlights the need for a more principled approach to measuring LLM judgment reliability.

A common approach for establishing reliability across LLM judgment models is the employment of inter-rater reliability \citep{kollitsch2024does, wang2024prompt}. This metric, as we will show, is an insufficient quantification of reliability as the metric does not take into account the additional complexity of the LLM judge being a model itself and containing randomness. Beyond inter-rater reliability, the LLM-as-a-judge literature predominantly relies upon accuracy and bias metrics to assess reliability \citep{chen2024humans, wei2024systematic, ye2024justice}. Some researchers advocate for combining responses from different LLM judgment models to improve reliability \citep{gu2024survey, patel2024aime}. To quantify reliability of LLM judgments, human judgments are often used as a benchmark, with methods relying upon agreement with human evaluation and correlation with human-annotated error scores \citep{jung2024trust, fu-etal-2023-large, dong2024can}. Unlike these approaches, which often rely on external benchmarks or human input, our work focuses on the internal consistency of LLM judgments, a critical aspect of reliability that has been largely overlooked in the LLM-as-a-judge paradigm. Our work pioneers the investigation of internal consistency reliability in this context, specifically by isolating the impact of random seed variation, addressing a significant gap in the literature.

\section{Limitations of Current Reliability Measures}

To illustrate the limitations of inter-rater reliability for LLM judgment reliability, we present a preliminary example. While inter-rater reliability, calculated as the percentage of agreement between raters and detailed further in \citep{hallgren2012computing}, is a useful metric in some contexts, it is not well-suited for the unique characteristics of LLMs as judges (specifically, the inherent randomness within LLM evaluators themselves). 

To demonstrate this, we performed an analysis using a simplified version of our Reliability Framework, which is described fully in Section \ref{sec:framework}. We obtained judgments from three LLM evaluators on responses to BIG-Bench Hard (BBH) questions from \citet{suzgun2022challenging}, using a temperature of 0.25. Each LLM provided 100 judgments for each question, varying only the random seed. We then calculated the inter-rater reliability across the three LLM evaluators for each replication of judgments (one set per replication), for a total of 100 inter-rater reliability values.

Table \ref{tab:irr} details the drastic variation in inter-rater reliability, ranging from 0.167 to 1.00. This disconcertingly wide range demonstrates that inter-rater reliability is highly sensitive to random seed variation, making it an unreliable indicator of true judgment quality. This instability in inter-rater reliability underscores the need for a more robust and appropriate measure of LLM judgment reliability, further motivating the development of our proposed (internal consistency) reliability framework.

\begin{table}[t]
    \centering
    \begin{tabular}{cccccc}
         & \textbf{Min} & \textbf{Q1} & \textbf{Q2} & \textbf{Q3} & \textbf{Max} \\
         \midrule
         \textbf{IRR} & 0.167 & 0.395 & 0.433 & 0.469 & 1.000\\
        \bottomrule
    \end{tabular}
    \caption{Variation of the inter-rater reliability (IRR) across 3 LLM evaluators across replications.}
    \label{tab:irr}
\end{table}

\section{Reliability Framework}
\label{sec:framework}

Our framework evaluates the reliability of the LLM-as-a-judge paradigm across diverse question formats and difficulty levels. Using LLM responses to benchmark questions, judgment LLMs are prompted repeatedly to select the "best" response based on factors like accuracy, utility, and relevance. The LLM is prompted for evaluation 100 times, varying only the replication while holding other factors constant. All judgment results are then assessed to uncover reliability by applying McDonald's omega.

The reliability framework begins by utilizing the widely accepted benchmarks BIG-Bench Hard (BBH) from \citet{suzgun2022challenging}, SQuAD from \citet{rajpurkar2016squad}, and MT-Bench from \citet{zheng2023judging}. These benchmarks offer two key advantages. First, they present diverse question formats: BBH and SQuAD questions are single-turn with specific target answers, while MT-Bench questions are multi-turn and open-ended. Second, they provide varying levels of difficulty, as evidenced by the significantly different accuracy rates of LLMs on BBH and SQuAD questions shown in Table \ref{tab:bench_acc}. This diversity allows us to assess judgments under a range of conditions.

One question from each category within each benchmark is randomly selected, for a total of 55 questions. Next, five responses for each question are generated using Chain-of-Thought prompting, as illustrated in Figure \ref{fig:prompt_resp}. The variation in prompting by benchmark is a result of the structuring of each benchmark; BBH responses require a single question, SQuAD responses require a single question as well as context surrounding the  question, and MT-Bench requires a two-turn dialogue.

\begin{figure}[h]
\centering
\begin{subfigure}[b]{0.49\textwidth}
\begin{tcolorbox}
Include step-by-step reasoning in answering the
following question: \\
\text{[Question]}
\end{tcolorbox}
\caption{Prompt template for generating responses to BBH questions.}
\label{fig:prompt_bbh}
\end{subfigure}
\hfill
\begin{subfigure}[b]{0.5\textwidth}
\begin{tcolorbox}
Include step-by-step reasoning in answering the following question: \\
Context: [Context] \\
Question: [Question] \\
Answer: 
\end{tcolorbox}
\caption{Prompt template for generating responses to SQuAD questions.}
\label{fig:prompt_squad}
\end{subfigure}
\begin{subfigure}[b]{0.5\textwidth}
\begin{tcolorbox}
Include step-by-step reasoning in answering the following question: \\
\text{[Question 1]} \\
\text{[Response 1]} \\
\text{[Question 2]}
\end{tcolorbox}
\caption{Prompt template for generating responses to the second turn and final of MT-Bench questions. To generate responses from the first turn, the template is identical to the BBH prompt.}
\label{fig:prompt_mtb}
\end{subfigure}
\caption{Prompt templates to generate responses from each of the varying benchmark types.}
\label{fig:prompt_resp}
\end{figure}

The model responses are then combined in a single prompt, as shown in Figure \ref{fig:judge_resp}. To prevent judgment bias, the five responses to the respective benchmark question are included in the prompt in a random order as response options [A] - [E]. This random order is held constant across judgment replications, but shuffled for each new benchmark question judgment prompt. In addition to the five responses, the judgment prompt includes the original benchmark question (and context or previous turns if relevant) and prompts the LLM for a judgment of best response including Chain-of-Thought. Here, Chain-of-Thought from \citet{wei2022chain} is again utilized as it has been shown to dramatically improve LLM performance \citet{feng2024towards, suzgun2022challenging, chu2023survey}. For the multi-turn responses from MT-Bench, options [A]-[E] are explicitly labeled as responses to the final turn, as shown in Figure \ref{fig:judge_mtb}.

\begin{figure}[]
\centering
\begin{subfigure}[b]{0.49\textwidth}
\begin{tcolorbox}
You are a fair and objective judge tasked with selecting the strongest of the following responses to the provided question. Base your judgment upon the accuracy, utility, and relevance of each. Do not consider length of response, positioning of response or title of response in your judgment. Output the letter of the best response followed by an explanation, and strictly follow the following format: "Best Response: [[letter]]". \\
Question: [Question] \\
Responses: \\
\text{[A]}: \text{[Response A]} \\
\text{[B]}: \text{[Response B]} \\
\text{[C]}: \text{[Response C]} \\
\text{[D]}: \text{[Response D]} \\
\text{[E]}: \text{[Response E]} \\
\end{tcolorbox}
\caption{Prompt template for generating judgments of responses to BBH and SQuAD questions.}
\label{fig:judge_bbh}
\end{subfigure}
\begin{subfigure}[b]{0.5\textwidth}
\begin{tcolorbox}
You are a fair and objective judge tasked with selecting the strongest of the following responses to the second provided question ("Question 2"). Question 1 is only provided for context. Base your judgment upon the accuracy, utility, and relevance of each. Do not consider length of response, positioning of response or title of response in your judgment. Output the letter of the best response followed by an explanation, and strictly follow the following format: "Best Response: [[letter]]". \\
Question 1: [Question 1] \\
Question 2: [Question 2] \\
Responses: \\
\text{[A]}: \text{[Response A]} \\
\text{[B]}: \text{[Response B]} \\
\text{[C]}: \text{[Response C]} \\
\text{[D]}: \text{[Response D]} \\
\text{[E]}: \text{[Response E]} \\
\end{tcolorbox}
\caption{Prompt template for generating judgments of responses to MT-Bench questions.}
\label{fig:judge_mtb}
\end{subfigure}
\caption{Prompt templates for generating LLM evaluations of responses for each of the varying benchmarks.}
\label{fig:judge_resp}
\end{figure}

Each LLM utilized for judgments is prompted 100 times, varying only the replication, to evaluate which of the responses to a given benchmark question is best (as defined in the prompt) out of the five responses provided. Given the employment of Chain-of-Thought in the initial responses to a given benchmark question, a single best response should exist in response options [A] - [E] for a judge to uncover because each of the five responses contains reasoning in addition to the question response. Similar logic is employed in \citet{zheng2023judging} in assessment of judgments as well.

The reliability of the judgments is then calculated using internal consistency reliability, a cornerstone of psychometric analysis and has been extensively studied. Cronbach's alpha, introduced by \citet{cronbach1951coefficient}, has been widely used to assess internal consistency, but relies on several assumptions that may not always hold \citep{malkewitz2023estimating, agbo2010cronbach}. As proposed by \citet{mcdonald2013test}, McDonald's omega addresses these limitations by accounting for a broader range of factor structures. Ultimately, omega provides a more robust measure of internal consistency and has been shown to be more robust to deviations from assumptions than alpha \citep{stensen2022internal}. In this work, we adopt McDonald's omega as our primary metric, recognizing its superior quantification of reliability. 

\subsection{McDonald's Omega}

McDonald's omega ($\omega$) is formulated as 

\begin{equation*}
    \dfrac{\left(\sum_{i=1}^{n}\lambda_i\right)^2}{\left(\sum_{i=1}^{n}\lambda_i\right)^2 + \sum_{i=1}^{n}\theta_{ii}+2\sum_{1\le i < j \le n}\theta_{ij}}
\end{equation*}

where $n$ is the number of replications which, in this setting, is the number of times the LLM makes a judgment. The parameter $\lambda_i$ is the factor loading of the i-th replication. This represents the strength of the relationship between the i-th judgment and the underlying "true" judgment. In simpler terms, factor loadings indicate how much each replication contributes to our understanding of the overall judgment being measured. A higher factor loading suggests that the replication is a strong indicator of the underlying true judgment. The error variance of the i-th replication is given by $\theta_{ii}$, which represents the variability or "noise" associated with the i-th judgment. Finally, the covariance between the errors of replications i and j is given by $\theta_{ij}$. This measures the extent to which the errors in different judgments are related. Essentially, McDonald's omega assesses how much of the observed variation in judgments reflects the true underlying value and how much is due to random error or inconsistencies in the measurement process.

To apply omega, we assume:

\begin{enumerate}
    \item Additive Measurement Error: Error in each LLM judgment replication is independent of the true judgment value. In simpler terms, the error does not systematically over- or underestimate the true judgment. This is a common psychometrics assumption, particularly when dealing with judgments that are expected to be relatively consistent \citep{alagumalai2005classical, wadkar2016assessing, liu2010impact}.
    \item Uncorrelated Errors: Errors in different LLM judgment replications are independent of each other. In other words, the error in one judgment does not influence the error in another judgment. This assumption is trivial in our setting, as each LLM judgment is generated independently so errors are guaranteed to be uncorrelated.
    \item Single Latent Trait: All LLM judgments are attempting to measure the same underlying "true" judgment. In our case, each LLM judgment replication aims to identify the most accurate, relevant, and useful response to the same prompt, aligning with this assumption.
\end{enumerate}

\section{Experiments and Data}
\label{sec:experiments}

Responses to benchmark questions, using the templates outlined in Figure \ref{fig:prompt_resp}, are generated from five highly popular open-source LLMs: LLaMA-3-8B, Vicuna-7B-v1.5, Gemma-7B, Phi-2, and Falcon-7b \citep{dubey2024llama, chiang2023vicuna, team2024gemma, javaheripi2023phi, falcon40b}. These models were chosen due to their popularity, with over 100,000 downloads in the prior month on Hugging Face.\footnote{As of November 2024 at \url{https://huggingface.co/models}} 
The generated responses are obtained using top-k of 50 and a temperature of 0.75.

Selecting models that are high performers in the Chatbot Arena \citep{chiang2024chatbot}, Starling-LM-7B-beta, Gemma-1.1-7b-it, and Meta-Llama-3-8B-Instruct are employed for the judgment task \citet{starling2023, gemma2024, llama3modelcard}. These selected models are all below the threshold of 10B parameters and are top performing models, ranked in the top 100, with rankings of 65 (Meta-Llama-3-8B-Instruct), 78 (Starling-LM-7B-beta), and 98 (Gemma-1.1-7b-it). In addition, we note that the selected models differ from those used for question answering in the prior step. This distinction is crucial to avoid circular reasoning, where LLMs would be judging responses they themselves generated. This approach also helps mitigate potential bias.

To further investigate the influence of fixed randomness on LLM outputs, we systematically varied the temperature parameter during the LLM evaluation process. Temperature, a key hyperparameter in LLMs, significantly impacts the randomness (or "novelty") of their outputs \citep{peeperkorn2024temperature, davis2024temperature}. We evaluated LLM judgments at five temperature levels: 1.0, 0.75, 0.5, 0.25, and 0 (machine epsilon), while keeping the top-k sampling parameter constant at 50. This systematic variation in temperature allowed us to observe how changes in the level of randomness within the LLM's generation process affect the variability and reliability of the subsequent judgments, providing a window into the impact of fixed randomness.

The calculation of McDonald's omega employs the \texttt{reliabiliPy} package \citep{rafael_valero_fernandez_2022_5830894}. It is important to note that in calculating the reliability, all judgments that did not provide a judgment of best response were treated as belonging to the same category ("Non Response") which was included as a category in addition to the five response categories "A" through "E". Consequently, the resulting calculated reliability serves as an upper bound on reliability because the variability in non-response judgments is ignored. This is further explored in the experiment results. Results are obtained using a pair of Quadro RTX 8000 GPUs with 48GB of memory each and the transformers library \citep{wolf2020transformers}.

\section{Experimental Results}

\begin{table}[!th] 
\begin{subtable}{.45\textwidth}
\centering
\begin{tabular}{cccc}
 \textbf{Temperature} & \textbf{BBH} & \textbf{SQuAD}  & \textbf{MT-Bench}\\
 \midrule
\textbf{1} & 0.702 & 0.632 & 0.462\\
\textbf{0.75} & 0.713 & 0.639 & 0.618\\
\textbf{0.5} & 0.703 & 0.644 & 0.476\\
\textbf{0.25}& 0.677 & 0.598 & 0.64\\
\textbf{0} & 1 & 1 & 1\\
\bottomrule
\end{tabular}
\caption{Omega reliability of Starling-LM-7B-beta judgments.}
\label{tab:starling_reliab} 
\end{subtable}

\vspace{1em} \begin{subtable}{.45\textwidth}
\centering
\begin{tabular}{cccc}
 \textbf{Temperature} & \textbf{BBH} & \textbf{SQuAD}  & \textbf{MT-Bench}\\
 \midrule
\textbf{1} & 0.712 & 0.632 & 0.59\\ 
\textbf{0.75} & 0.698 & 0.655 & 0.556 \\
\textbf{0.5} & 0.680 & 0.554 & 0.602\\ 
\textbf{0.25}& 0.661 & 0.533 & 0.421 \\ 
\textbf{0} & 1 & 1 & 1\\
\bottomrule
\end{tabular}
\caption{Omega reliability of Meta-Llama-3-8B-Instruct judgments.}
\label{tab:llama_reliab} 
\end{subtable}
\begin{subtable}{.45\textwidth}
\centering
\begin{tabular}{cccc}
   \textbf{Temperature} & \textbf{BBH} & \textbf{SQuAD}  & \textbf{MT-Bench}\\
 \midrule
\textbf{1} & 0.723 & 0.64 & 0.605 \\
\textbf{0.75} & 0.77 & 0.73 & 0.585\\
\textbf{0.5} & 0.788 & 0.751 & 0.732 \\ 
\textbf{0.25}& 0.803 & 0.77 & 0.637 \\
\textbf{0} & 1 & 1 & 1\\
\bottomrule
\end{tabular}
\caption{Omega reliability of Gemma-1.1-7b-it judgments.}
\label{tab:gemma_reliab}
\end{subtable}
\caption{Omega reliability by LLM judge across temperatures and benchmarks.}
\label{tab:exp_judg_reliab}
\end{table}

Our findings reveal several key insights. First, we observed that the ideal temperature for reliable LLM judgment is not universal; it varies depending on the specific LLM and the task, as revealed in Table \ref{tab:exp_judg_reliab}. For example, some models (like Gemma-1.1-7b-it on SQuAD) exhibit decreased reliability at higher temperatures, while others (like Meta-Llama-3-8B-Instruct on BBH) show the opposite trend. Consequently, careful temperature tuning is crucial for each LLM and task to balance output novelty and judgment reliability. The ability to maintain reliability while increasing temperature, however, is a promising avenue for future research, as it provides opportunities to explore methods that generate novel, reliable, and high-quality outputs.

\begin{table}[!th]
 \centering
\begin{tabular}{cc}
    \textbf{Reliability} & \textbf{Interpretation} \\
   \hline
    $\mathbf{\boldsymbol{\alpha} \boldsymbol{>} 0.9}$ & Excellent\\
    $\mathbf{0.9 > \boldsymbol{\alpha} > 0.8}$ & Good \\
    $\mathbf{0.8 > \boldsymbol{\alpha} > 0.7}$ & Acceptable \\
    $\mathbf{0.7 > \boldsymbol{\alpha} > 0.6}$ & Questionable \\
    $\mathbf{0.6 > \boldsymbol{\alpha} > 0.5}$ & Poor \\
    $\boldsymbol{\alpha} \mathbf{< 0.5}$ & Unacceptable \\
    \hline
\end{tabular}
\caption{Rule of thumb for interpreting reliability measures.}
\label{tab:rule_thumb}
\end{table}

Furthermore, the results in Table \ref{tab:exp_judg_reliab} highlight that LLM judgment reliability is a major concern. It's important to note that the calculated reliability values are likely optimistic estimates of the true reliability, as judgment outputs that did not select a response were treated as belonging to the same category ('Non-response'). Despite this, the overall reliability scores are troubling. Adhering to the widely accepted reliability rule of thumb (Table \ref{tab:rule_thumb}), Starling-LM-7B-beta and Meta-Llama-3-8B-Instruct exhibited questionable reliability across benchmarks, with SQuAD and MT-Bench results consistently falling below the acceptable range (Tables \ref{tab:starling_reliab} and \ref{tab:llama_reliab}). This raises serious questions about the trustworthiness of their judgments.

Moreover, our results suggest a potential performance-reliability trade-off. The ranking of LLM judgment reliability (Gemma-1.1-7b-it > Starling-LM-7B-beta > Meta-Llama-3-8B-Instruct) inverts their Chatbot Arena performance ranking. This suggests that models optimized for performance may sacrifice reliability in their evaluations.

\begin{table}[t]
    \centering
    \begin{tabular}{lccccc}
          & \textbf{Min} & \textbf{Q1} & \textbf{Q2} & \textbf{Q3} & \textbf{Max} \\
         \midrule
         \textbf{BBH} & 0 & 0 & 0 & 1 & 2 \\
         \textbf{SQuAD} & 3 & 5 & 5 & 5 & 5 \\
         \bottomrule
    \end{tabular}
    \caption{Five-number summary of accuracies of responses [A]-[E] across all sampled questions for each benchmark.}
    \label{tab:bench_acc}
\end{table}

Finally, we find that benchmark quality impacts judgment reliability. SQuAD, despite its higher overall answer accuracy (Table \ref{tab:bench_acc}), paradoxically exhibits lower judgment reliability than BBH (Table \ref{tab:exp_judg_reliab}). We attribute this to the difficulty in distinguishing the best answer among multiple highly accurate responses in SQuAD. This highlights a critical weakness of current LLM judges: struggling with nuanced distinctions between correct answers.

\section{Application}

We explore the influence of LLM reliability on practical applications, focusing on how LLMs function as judges within the Head-to-Tail benchmark \citep{sun2023head}. This benchmark assesses LLM knowledge by posing questions about entities of varying popularity ("head," "torso," "tail"). We replicated the benchmark's "head" question set for the Academic domain using Vicuna-7B, employing Starling-LM-7B-beta, Gemma-1.1-7b-it, and Meta-Llama-3-8B-Instruct as judges. Our goal is to demonstrate the strong connection between high reliability and consistent evaluation.

We replicated the $A_{LM}$ metric (accuracy judged by an LLM) reported in \citet{sun2023head} using Vicuna-7B. Table \ref{tab:acc_alm} shows close agreement between our replicated $A_{LM}$ values and the original results, confirming the reliability and reproducibility of LLM-based evaluations in this structured task. The consistent performance across different LLM judges further supports this.

Critically, our analysis reveals a strong link between high reliability and stable evaluation. Table \ref{tab:app_reliab} shows remarkably high reliability scores for our LLM judges across Head-to-Tail replications. Similarly, it is clear from the standard errors in the $A_{LM}$ estimates in Table \ref{tab:acc_alm} that minimal variability exists across replications. This demonstrates that highly reliable LLM judges produce consistent and trustworthy evaluations, further supporting the validity of reliability.

\begin{table}
    \centering
    \begin{tabular}{cc}
        \textbf{Judge} & $\mathbf{A_{LM}}$ \textbf{Estimate} \\
        \hline
        \textbf{Sun et al., 2024} & 0.039\\
        & (N/A)\\
         \textbf{Starling-LM-7B-beta} & 0.0396 \\
         & (0.00037)\\
         \textbf{Meta-Llama-3-8B-Instruct} & 0.0395 \\
         & (0.00036)\\
         \textbf{Gemma-1.1-7b-it} & 0.0395 \\
         & (0.00033)\\
         \hline
    \end{tabular}
    \caption{Average $A_{LM}$ of Vicuna-7B responses to Head questions across replications of judgment models as compared to the results reported by Sun et al., 2024. Corresponding standard errors are provided beneath respective estimates for our judgment models.}
    \label{tab:acc_alm}
\end{table}

\begin{table}
    \centering
    \begin{tabular}{cc}
         \textbf{Omega} & \textbf{Vicuna-7B} \\
         \hline
         \textbf{Starling-LM-7B-beta} & 0.9901 \\
         \textbf{Meta-Llama-3-8B-Instruct} & 0.9896 \\
         \textbf{Gemma-1.1-7b-it} & 0.9893 \\
         \hline
    \end{tabular}
    \caption{Omega reliability of our judgment models Starling-LM-7B-beta, Gemma-1.1-7b-it, and Meta-Llama-3-8B-Instruct across replications for Head question responses by Vicuna-7B.}
    \label{tab:app_reliab}
\end{table}

It is unsurprising that the Head-to-Tail reliability is much higher than that of our experiments given that the LLM judgments in the Head-to-Tail paradigm only have two options ('correct' or 'incorrect') and are provided with a ground truth in the prompt. Thus, this employment of LLM-as-a-judge is much more simplistic than our experiments and would be expected to show a higher reliability. While Head-to-Tail uses a simplified paradigm, it offers valuable insight into the reliability-consistency relationship. The observed high reliability, coupled with minimal variation in $A_{LM}$, strongly suggests that with careful design and clear criteria, LLMs can be dependable judges. Future research should extend these findings to more complex scenarios, developing methods to structure complex judgment tasks to leverage LLM reliability and enhance trustworthiness in diverse applications. Our findings reinforce reliability's importance in trustworthy LLM evaluation and highlight the potential for reliable LLM judgment in well-defined tasks.

\section{Conclusions}

Current LLM-as-a-judge methods, relying on single outputs, mask inherent judgment variability, creating a false sense of reliability \citep{ouyang2023llm, wei2024systematic}. This "fixed randomness" (Section \ref{sec:intro}), poses risks, especially in high-stakes applications like medical AI evaluation. We address this by introducing a novel framework for evaluating LLM-as-a-judge reliability, focusing on internal consistency across replicated evaluations with varying random seeds — a crucial, previously overlooked aspect. 

Our results validate our framework's importance. LLM judgment reliability is a significant concern, with several models exhibiting questionable reliability across benchmarks like SQuAD and MT-Bench. SQuAD's low reliability, despite high accuracy, reveals LLMs' difficulty with nuanced distinctions. Our research also suggests a performance-reliability trade-off, which our framework helps navigate. Finally, Head-to-Tail results further demonstrate our framework's value, as high reliability scores and minimal variability highlight the link between reliability and consistent evaluation.

This work provides a robust framework for assessing LLM-as-a-judge reliability, addressing a critical gap and developing a crucial tool for building trust in LLM-powered systems. As LLMs are increasingly used in complex applications, reliably assessing their judgment is paramount. Our framework empowers informed decisions about LLM deployment, promoting responsible usage. Prioritizing reliability is key to unlocking LLMs' potential while ensuring their ethical use.

\section{Limitations}

While this work represents a significant step forward in understanding and assessing the reliability of LLMs, it is important to acknowledge its limitations. Our work is currently focused on single and multi-turn benchmarks. These benchmarks were selected to provide a robust and generalizable understanding of the impact of reliability under varying and widely used contexts, including both structured and open-ended tasks. Future research can extend our approach to additional datasets and benchmarks. Further, developing domain-specific practical guidelines for interpreting and applying reliability metrics is an important area for future work given that different domains have varying tolerances for variability. For instance, fields like advertising may be more tolerant of variability, while fields like medicine require much higher levels of certainty. By providing domain-specific guidance, we can ensure that LLMs are used responsibly and effectively in various applications.

Despite these limitations, our work provides a strong foundation for future research on LLM reliability. By introducing a rigorous framework for evaluating LLM-as-a-judge responses, we offer valuable insights into the impact of reliability of LLM outputs and provide a roadmap for improving the quality and trustworthiness of LLMs.

\section{Ethics Statement}

This research does not raise any ethical concerns based on the theories or datasets employed. The benchmarks (BBH, SQuAD, MT-Bench, and Head-to-Tail) are publicly available and utilize publicly available data sources. As such, no privacy violations are anticipated. However, researchers should exercise caution when applying this work to proprietary datasets, as these may involve specific privacy regulations and ethical considerations.

\bibliography{custom}




\end{document}